# The Effects of Image Pre- and Post-Processing, Wavelet Decomposition, and Local Binary Patterns on U-Nets for Skin Lesion Segmentation

Sara Ross-Howe, H.R. Tizhoosh

*Kimia Lab, University of Waterloo,* Waterloo, Ontario, Canada
{Sara.Ross,Tizhoosh}@uwaterloo.ca; http://kimia.uwaterloo.ca/

*Abstract*—Skin cancer is a widespread, global, and potentially deadly disease, which over the last three decades has afflicted more lives in the USA than all other forms of cancer combined. There have been a lot of promising recent works utilizing deep network architectures, such as FCNs, U-Nets, and ResNets, for developing automated skin lesion segmentation. This paper investigates various pre- and post-processing techniques for improving the performance of U-Nets as measured by the Jaccard Index. The dataset provided as part of the "2017 ISBI Challenges on Skin Lesion Analysis Towards Melanoma Detection" was used for this evaluation and the performance of the finalist competitors was the standard for comparison. The pre-processing techniques employed in the proposed system included contrast enhancement, artifact removal, and vignette correction. More advanced image transformations, such as local binary patterns and wavelet decomposition, were also employed to augment the raw grayscale images used as network input features. While the performance of the proposed system fell short of the winners of the challenge, it was determined that using wavelet decomposition as an early transformation step improved the overall performance of the system over pre- and post-processing steps alone.

## I. INTRODUCTION

Each year in the USA, over 5.4 million cases of Non-Melanoma skin cancer are treated in more than 3.3 million people [1]. Over the past three decades, more people have had skin cancer than all other cancers combined [3]. The risk of skin cancer also increases with age, and between 40 and 50 percent of Americans who live to the age of 65 will encounter at least one episode of Basal Cell Carcinoma or Squamous Cell Carcinoma in their lifetime [5]. And tragically, one person dies of Melanoma every 54 minutes in the USA [2].

There are three main types of skin cancer, namely: Basal Cell Carcinoma, Squamous Cell Carcinoma, and Melanoma. Basal Cell Carcinoma is the most common form of skin cancer, with over 4 million cases treated each year in the USA [1]. Squamous Cell Carcinoma is the second most common form of skin cancer, with 1 million cases treated in the USA each year [1]. Melanoma accounts for <1% of all cases of skin cancer, but contributes to the largest number of deaths [2].

Diagnosis of skin cancer begins with an examination of the skin lesion by a dermatologist, and this procedure is enhanced using a dermoscope. Dermoscopy images captured during the inspection are characterized by their high resolution and increased magnification (typically > 10x). The field of view has focused illumination and polarized light, which allows visualization of deeper skin structures and helps to cancel out surface reflections. In a clinical setting, dermoscopes often have fixed mountings, with durable chassis to achieve regulatory compliance. Recent accessories, such as the HandyScope [34], allow smartphones to operate as portable digital dermoscopes. These devices could help early diagnosis of skin cancer in the home healthcare environment, and enable lesions to be more frequently monitored and changes tracked over time.

Characteristics of malignant skin lesions can be described by the following defining features: Asymmetry, Border, Colour, Diameter, and Evolving [35]. Benign lesions are symmetrical, with smooth and even boards, consistent colour (often a single shade of brown) and are constant over time. While malignant skin lesions are asymmetrical, with uneven borders (can be scalloped or notched), contain a number of different shades (can be brown, tan, black, red, white, or blue), have larger diameters (often >6mm), and change over time (e.g. size, colour, shape, elevation, itching, crusting, or bleeding). Skin lesion segmentation is an important initial analysis step to be able to quantify these features. Segmentation has been demonstrated to improve classification accuracy [17], support treatment planning, and allow changes that occur to lesion boundaries to be tracked over time.

## II. BACKGROUND REVIEW

Deep learning is a recent and disruptive specialization of machine learning, which utilizes representation learning to organize and automatically extract progressive layers of features abstraction directly from raw data [6]. Deep learning networks have been demonstrated as very effective techniques for semantic segmentation in applications for machine vision [7]. Convolutional Networks (ConvNets) provide a foundation from which many more complex network architectures have been based for both image classification and segmentation tasks [7,8,9]. For example, Fully Connected Networks (FCNs) were one of the first network architectures to utilize ConvNets for end-to-end, pixel-to-pixel semantic segmentation [9]. FCNs utilized a convolution-deconvolution architecture and does not contain any fully connected layers.

AlexNet was introduced at the ImageNet 2012 competition, where it produced a top five test error of 15.4% on the difficult

ImageNet dataset, consisting of over 15 million images and representing over 22000 categories (the next best competitor achieved a 26.2% top 5 error rate) [7]. AlexNet consisted of 5 convolutional and 3 fully connected layers AlexNet utilized important concepts such as using ReLU as an activation function, and data augmentation and dropout to help manage overfitting [7].

Deep residual networks (ResNets) were introduced by the Microsoft Research Group as part of the ImageNet and COCO competitions in 2015, where they garnered first place in the categories of ImageNet detection, ImageNet localization, COCO detection, and COCO segmentation [10]. These ResNets were deeper than previous networks with up to 152 layers and offered a solution to the degradation problem that occurs in deep network structures, where a saturation and then rapid decrease in accuracy occurs during training. These degradations are not due to overfitting in the model, but rather are attributed to vanishing gradients. To address degradation problems in deep networks, ResNets utilizes shortcuts where outputs from previous layers skip one or more stacked layers and bring image content directly to deeper layers.

A special convolution-deconvolution architecture, referred to as U-Nets, was introduced as part of the ISBI cell tracking challenge in 2015 [11]. These U-Nets address special challenges with medical image segmentation, such as limited numbers of labelled training and testing cases, and the importance of precision in boundary definition. The U-Net architecture contains a contracting path where image features are extracted and a subsequent expanding path that supports localization of the segmentation borders. The U-Nets were demonstrated as successful segmentation methods against a number of biomedical tasks, including segmentation of neuronal structures in electron microscope stacks, and cell segmentation in light microscopic images.

The International Skin Imaging Collaboration (ISIC) is a partnership between industry and academia to promote the development of digital skin imaging applications and to help to reduce the global mortality rate from Melanoma. This consortium proposes standards for dermatologic imaging to improve quality and usefulness, and publishes public datasets with gold standard labels for the validation and development of diagnostic algorithms. On December 10, 2017, the ISIC launched the "2017 ISBI Challenges on Skin Lesion Analysis Towards Melanoma Detection", as an open competition to accelerate technology development in three areas of investigation: Lesion segmentation, detection and localization of visual dermoscopic features/patterns, and disease classification [20].

In the concluding results from the 2017 ISBI Challenge, the top five competing entries utilized deep network architectures with either FCNs, U-Nets, or ResNets [12,13,14,15,16]. The top results were obtained by a team at Mount Sinai, where they trained an FCN with 29 layers that used both RGB and HSV channels as inputs, achieving an official Jaccard Index of 0.765 in the competition [12]. U-Nets were utilized by the second-place entry with 3 down sampling layers and reflected up-sampling layers, and a fully connected layer at the bottom [13]. This entry utilized data augmentation to produce 20,000 training images and achieved an official Jaccard Index of 0.762 in the challenge.

III. DATASET

The dataset provided as part of the lesion segmentation section of the 2017 ISBI Challenge, consisted of dermoscopy images with a ground truth segmentation mask annotated by an expert clinician. The annotation process was done with either a semi-automated process (a user-provided seed point, a user-tuned flood-fill algorithm, and morphological filtering) or a manual process (from a series of user-provided polyline points) [20]. The lesion segmentation dataset consisted of 2000 training images, 150 validation images, and 600 testing images. To augment the data available for training deep networks, 90° rotation, 180° rotation, 270° rotation, x-axis mirror, and y-axis mirror transformations were applied to the original image (see Figure 1). The validation images were combined with the original training images to produce a total of 12900 training images and 3600 testing images.

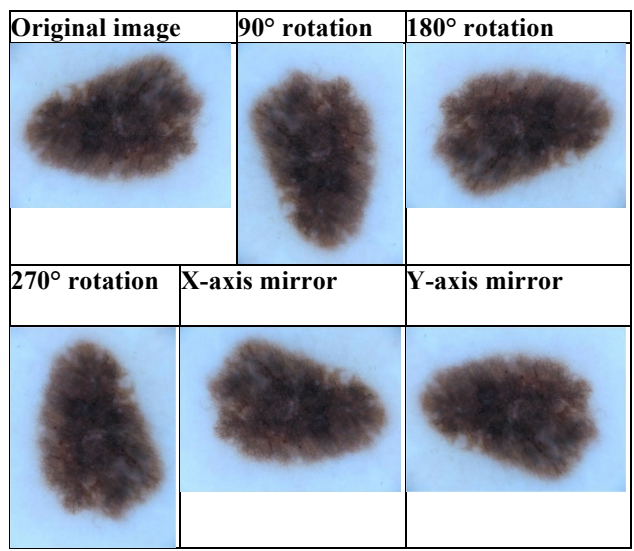

*Figure 1: Data Augmentation: Original image, 90° rotation, 180° rotation, 270° rotation, x-axis mirror, y-axis mirror*

There are many challenges with skin lesion segmentation, as demonstrated within the 2017 ISBI Challenge training dataset. Dermoscopy images can include dark hairs that occlude the lesion boundaries, and lesions can exist anywhere on the body, which can challenge visibility. Vignettes around strongly illuminated fields of view provide difficult shadow regions, and annotation marks used to establish image scale can confuse object detection algorithms. These images can also be of very low contrast, and the variety of skin pigmentations in a properly represented global population also adds to the complexity of the analysis. Examples of these challenges have been illustrated in Figure 2.

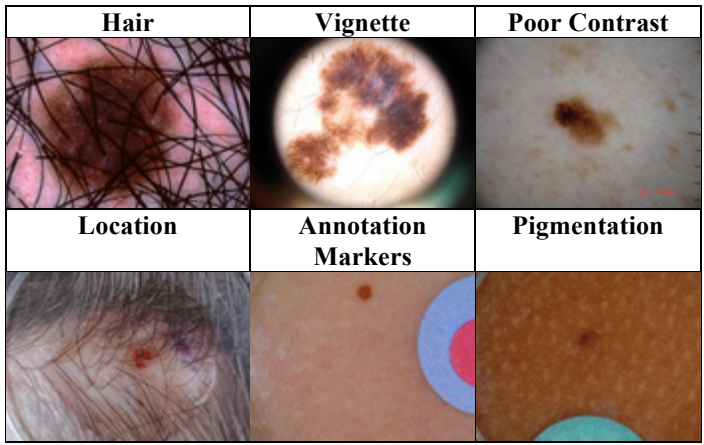

*Figure 2: Challenges in the 2017 ISBI Challenges on Skin Lesion Analysis Towards Melanoma Detection dataset*

IV. PROPOSED SOLUTION

The architecture of the proposed solution includes five core processing stages, which can be defined by image preparation, image pre-processing, image transformations, U-Net, and image post processing. In Figure 4, the stages of processing for the proposed system have been sequentially outlined, and the combined outputs of the proposed analysis at each step are illustrated.

*A. Image Preparation*

The first step of the outlined solution is to prepare the dermoscopy images for analysis. These source images are variable in size, are high resolution, and have 24-bit colour depth. The images are first resized to a more computationally manageable 216x216 dimensions. A 20-pixel border is then added around the image to buffer the outer image content, as the U-Net processing crops these external regions due to the loss of border pixels in every convolution. The image is then converted to grayscale to further reduce the image dimensions, and to match the format of images used in the original U-Net reference [11].

*B. Image Pre-Processing*

Segmentation techniques that were developed prior to the introduction of deep convolution-deconvolution networks relied on many pre-processing techniques. Important pre-processing steps included color space transformations, contrast enhancement, appropriate lesion localization, and artifact removal [21]. Common artifacts in dermoscopy images include: vignette frames, ink markings, scale rulers, skin lines, blood vessels, and hairs [21]. The DullRazor method is a commonly cited hair remove algorithm, which uses morphological closing to detect hairs and then removes them using bilinear interpolation [22]. To remove the vignette effect Wang et al, [23] utilize a three-step process that begins with isolating the image in the red coordinate of RGB. Concentric circular regions, with a tunable diameter, are first defined. The analysis process is then started at the center of the image. In the last step, the brightness of each concentric region is adjusted so that the average intensity is the same as the center region [23].

The first stage of pre-processing employed by the proposed system was to perform contrast enhancement. This ensures the images have consistent contrast between neighbouring areas and the region of interest. Contrast enhancement was accomplished by first calculating the histogram of the grayscale image. The top and bottom 2% of histogram intensities were then selected and used as cut off values. The histogram was then stretched to remap the darkest pixel to 0 and lightest pixel to 255 against the selected cut off thresholds.

The second step performed for pre-processing was for hair removal. Hairs were first identified by using an edge detection with 1-pixel silhouette. The method used for edge detection was the Python ImageFilter FIND_EDGES in the Python Imaging Library (PIL) [24]. Filtered output is then superimposed on the original image to reduce the identified line width by 2 pixels. As the lesions are defined by their darker colours, the edges detected were just filled in with white (instead of a blurring method with the neighboring skin colour).

The final stage in the image pre-processing procedure was to remove the defined vignette frame found in some images. These dark frames are caused from the strong illumination at the center of view because of the lighting from the dermoscope, or from the restricted diameter of the cannula on the dermoscope tip. The pseudocode for the vignette frame removal algorithm is seen below in Figure 3.

```
For up to 10 iterations
    Create a circular mask with
        radius = half the height of the input image –
        20 pixels + iteration*5 pixels
    Calculate the mean of the outer circle
        If the mean pixel value of the outer ring <
        6.0 then
            Fill the outer ring with the mean
            pixel value of the center region
            Return the corrected image
```

*Figure 3: Pseudocode for vignette removal algorithm*

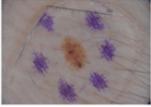

*Figure 4: Five core processing stages of the proposed solution*

The end result of the three combined pre-processing steps is illustrated in Figure 5 below.

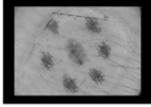

*Figure 5: Pre-processing steps in the proposed system*

### C. Image Transformations

A number of more advanced image transformation techniques were considered in an effort to have the U-Net converge faster and with fewer training images. The first image transformation technique investigated was Local Binary Patterns (LBP). LBP was first proposed by Ojala et al. [25] as a means to capture important textural information in an image. The LBP algorithm uses the value of the center pixel as a threshold in a neighbouring mask. The output from the thresholding process is then translated to a binary value. This technique has been used to define features for facial recognition [26] and medical image retrieval [27].

Wavelet transforms are another image transformation approach that has been important for image compression [28] and de-noising techniques [29]. The wavelet transform is time-frequency method where the optimal frequency band is adaptively determined to provide the best time-frequency resolution. The effects of applying the wavelet transform iteratively is to construct a multi-residual pyramidal representation of the image. In each subsequent layer of representation, the horizontal, vertical, and diagonal residuals are stored along with the approximation image. For the application of wavelet transforms for the proposed approach, the PyWavelet library [36] was used with the Daubechies 1 mother wavelet and three levels of decomposition were performed. The approximation image at each sub-band was resized to 216x216 images with a 20-pixel boundary added. See a sample of one level of decomposition illustrated in Figure 6.

### D. U-Net

The U-Net architecture, as defined by Ronneberger et al. [11] for biomedical image segmentation, was utilized for the proposed system. The U-Net was configured for 3 layers, with 64 filters in the initial convolution bank, a convolution filter size of 3x3, and an average pooling filter size of 2x2. The architecture of the U-Net used for investigations is illustrated in figure 7. The Python library used for the U-Net was adapted from the code source provided by Akeret et al. [30]. An Adam optimizer [31] was used with a fixed learning rate of 0.0001, and batch normalization was added to improve the accuracy of the system and enable faster convergence [32]. The loss function used a mean cross-entropy measure and the system was trained for 20 epochs.

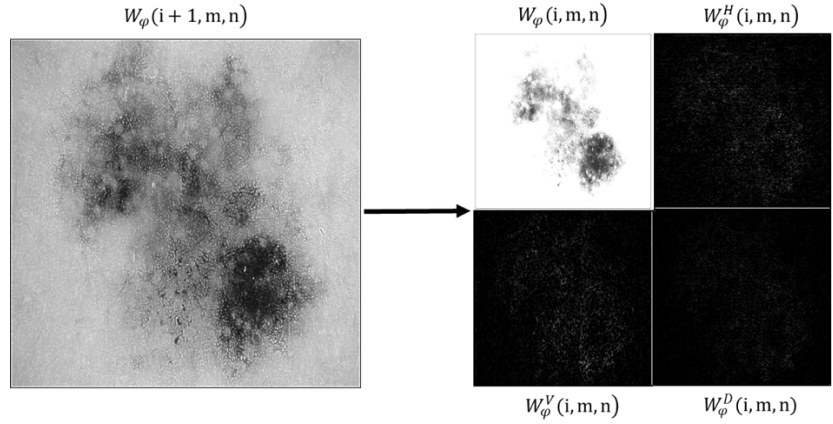

*Figure 6: Wavelet pyramid representation of dermoscopy image*

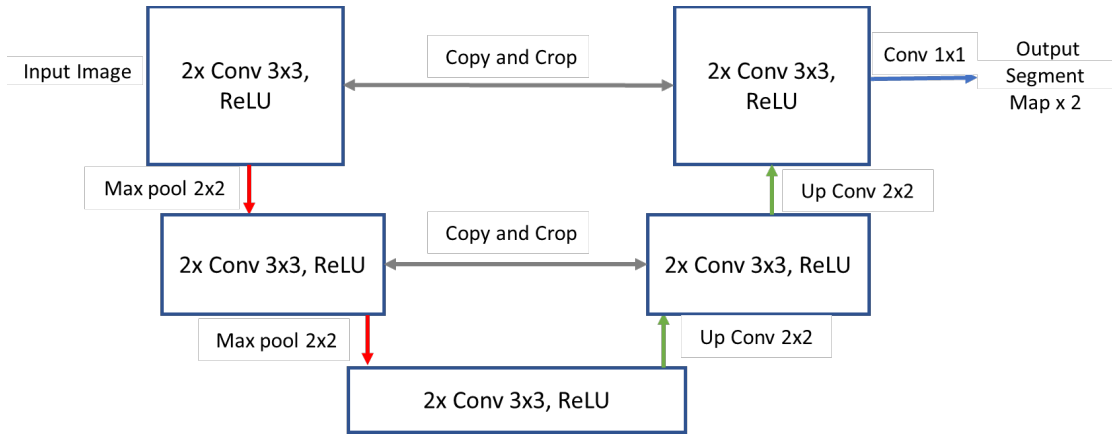

*Figure 7: U-Net architecture for skin lesion segmentation*

*E. Image Post-Processing*

The output from the U-Net are two pixel-by-pixel reconstructions of the segmentation mask. It was noted during experimentation that one of the masks represented more of the noise content in the image and the other the true lesion segment. Thus, instead of a binary argmax between the maximum of the two outputs, a static threshold cut off level of 0.5 was used against the better of the two images.

The binary segmentation mask that is produced by the binary thresholding, often contained regions representing noise artifacts. To remove these competing objects, and identify just the true single lesion object, a post-processing algorithm was applied. This algorithm leveraged on the intuition that the skin lesions were typically centered in the image and were one of the largest objects in the frame of view. The pseudocode for this algorithm is listed below in Figure 8.

V. EXPERIMENTS AND RESULTS

The quality metric used to assess the performance of the training and testing output was the Jaccard Index:

$$J(A, B) = \left|\frac{A \cap B}{A \cup B}\right|.$$

This metric was the gold standard used to rank the performance of submissions for the 2017 ISBI Challenges on Skin Lesion Analysis Towards Melanoma Detection.

Each of the 20 training epochs consisted of 32 iterations (or segments) with batches of 16 randomly selected images. Testing was performed against 72 randomly selected batches from the testing set that consisted of 16 images in each. The Jaccard Index reported represents the last epoch in training. The mean and standard deviation of the Jaccard Index for all 72 testing batches is also presented as assessment criteria.

> *Find the objects in the binary image*
>     *Select the 3 objects with the largest surface area*
>     *Locate the centroid of these objects*
>         *Determine the distance of the centroid to the nearest bounding edge.*
>         *Select the object that is the furthest distance from an edge and return this mask*

*Figure 8: Pseudo code for post-processing in proposed system*

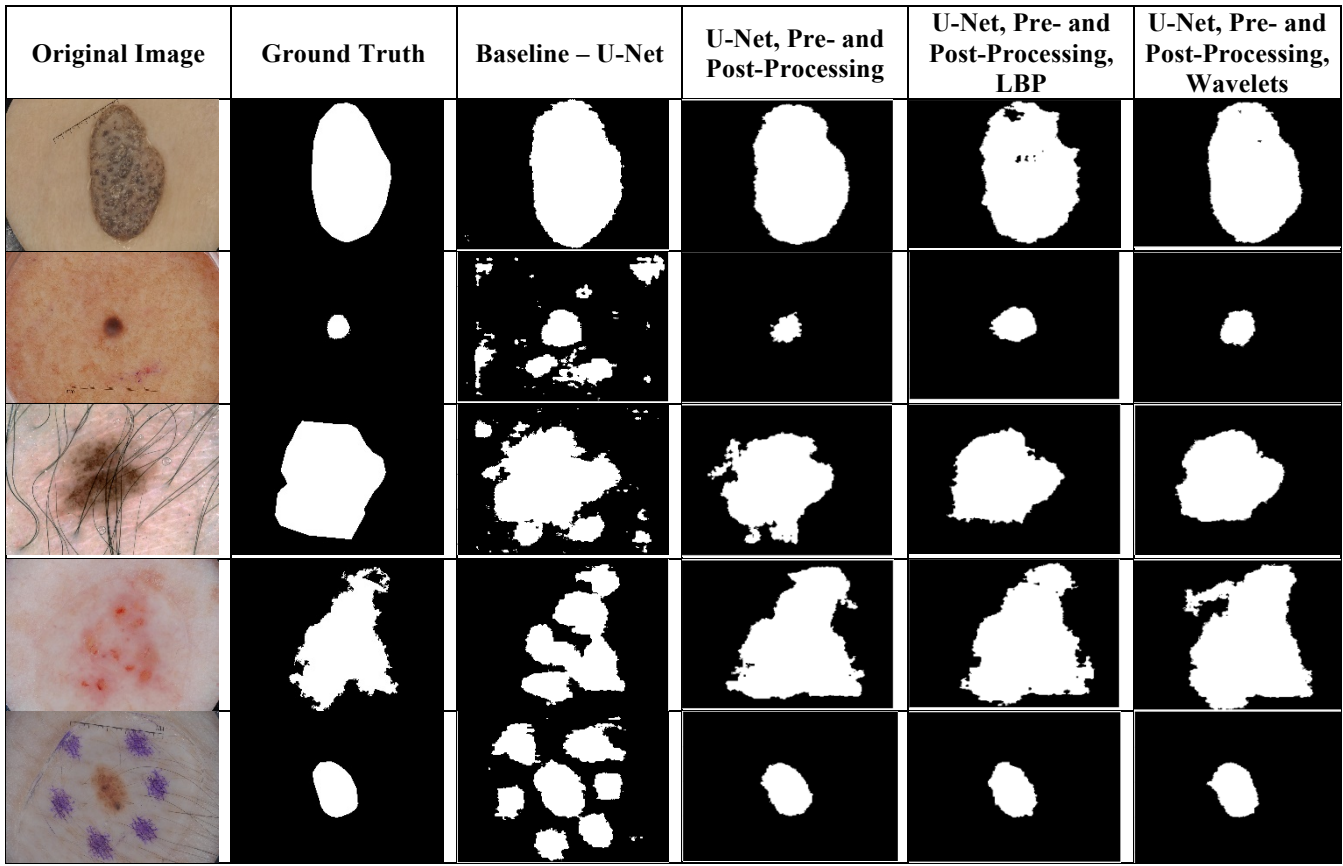

*Figure 9: Examples of processed dermoscopy images by the proposed system*

|              | A     | B     | C     | D     |
|--------------|-------|-------|-------|-------|
| Training     | 51.62 | 61.97 | 59.70 | 62.85 |
| Testing μ    | 51.13 | 57.17 | 56.20 | 59.73 |
| Testing σ    | 0.08  | 0.07  | 0.08  | 0.07  |

*Table 1: Training and testing results for the proposed system: A) Baseline U-Net, B) U-Net plus pre- and post-processing, C) U-Net plus pre- and post-processing plus LBP, D) U-Net plus pre- and post-processing plus wavelets*

Computations were run on an Amazon Web Service virtual instance, configured as a Windows Server 2016 machine with 4 CPUs, 16 GB of default memory, and a single Tesla K80 GPU. Tensorflow 1.4.1 was used with GPU processing enabled. The computation time to complete 20 training epochs for each proposed scenario was approximately 6 hours. The final training and testing results are presented in Figure 9 and in Table 1 above. The assessment criteria used to compare the performance of the different configurations was the Jaccard Index as a percentage calculated from the ground truth. Individual evaluations are given in Table 1 for five different scenarios, namely: 1) Baseline U-Net (raw grayscale input images with no pre-processing), 2) U-Net with pre- and post-processing, 3) U-Net with pre- and post-pressing and an LBP transformation, and 3) U-Net with pre- and post-processing and three layers of wavelet decomposition.

From the results in Table 1, it can be seen that the pre- and post-processing improved the U-Net performance on the testing set by an increase of 6.04% points above that of the baseline configuration. Adding the LBP to the network input produced a degradation in system performance, as seen in column 4 of Table 1. However, the use of wavelet decomposition improved the overall performance of the network by an additional 2.56% points above the scenario where just the pre- and post-processing techniques were used. The best performance was achieved when using pre- and post-processing in conjunction with the three layers of wavelet decomposition. The average Jaccard Index on the testing folds for the best configuration was 59.73%, which would translate to a ranking of 18 out of 20 on the final ranking leader board of the 2017 ISBI Challenge.

In Figure 9, the output of the different proposed configurations is illustrated using five different example images. These images represent more challenging cases and include poor contrast, small lesions, marker annotations, hair, and faint lesions.

## VI. CONCLUSIONS AND FUTURE WORK

In conclusion, pre- and post-processing of dermoscopy images improved the performance of U-Nets for skin lesion segmentation. When local binary patterns were extracted from the processed dermoscopy images and used as additional inputs into the U-Net, weaker performance was demonstrated than when pre- and post-processing techniques were used in isolation. Wavelet decompositions of the pre-processed image did provide additional performance improvements over pre- and post-processing of inputs and outputs alone. This is hypothesized to be due to the feature extraction and de-noising benefits of the wavelet transformation.

Future work would include making additional improvements to both the pre- and post-processing techniques, as they demonstrated direct positive impact on the resultant Jaccard Index. Specific areas of focus would include enhancements to both the hair removal and vignette correction pre-processing steps.

The U-Net training and testing for this research was conducted with constrained computational resources, but the effect of running with much larger datasets, increased network layers, and over more epochs should be investigated further. It would be beneficial to determine if the use of wavelet decomposition could improve on the accuracy achieved by Yading et al. [12], and if the technique translates to improvements for many different kinds of image data and ConvNet architectures.